\pdfoutput=1
\documentclass[10pt,twocolumn,letterpaper]{article}

\usepackage{iccv}
\usepackage{times}
\usepackage{epsfig}
\usepackage{graphicx}
\usepackage{amsmath}
\usepackage{amssymb}
\usepackage{mathrsfs}
\usepackage{multirow}
\usepackage{subfig}
\usepackage{booktabs}

\newcommand*{\affaddr}[1]{#1} 

\newcommand*{\email}[1]{\texttt{#1}}

\usepackage[pagebackref=true,breaklinks=true,letterpaper=true,colorlinks,bookmarks=false]{hyperref}

\iccvfinalcopy 

\def\iccvPaperID{10665} 

\ificcvfinal\fi

\begin{document}

\title{MetaMixer: A Regularization Strategy for Online Knowledge Distillation}

\author{%
Maorong Wang, Ling Xiao, Toshihiko Yamasaki\\
\affaddr{The University of Tokyo}\\
\email{\tt\small\{ma\_wang,ling,yamasaki\}@cvm.t.u-tokyo.ac.jp}
}

\maketitle
\ificcvfinal\thispagestyle{plain}\fi

\begin{abstract}
   Online knowledge distillation (KD) has received increasing attention in recent years. However, while most existing online KD methods focus on developing complicated model structures and training strategies to improve the distillation of high-level knowledge like probability distribution, the effects of the multi-level knowledge in the online KD are greatly overlooked, especially the low-level knowledge. Thus, to provide a novel viewpoint to online KD, we propose MetaMixer, a regularization strategy that can strengthen the distillation by combining the low-level knowledge that impacts the localization capability of the networks, and high-level knowledge that focuses on the whole image. Experiments under different conditions show that MetaMixer can achieve significant performance gains over state-of-the-art methods. 
\end{abstract}


\begin{figure*}[h]
\begin{center}
   \includegraphics[width=0.95\linewidth]{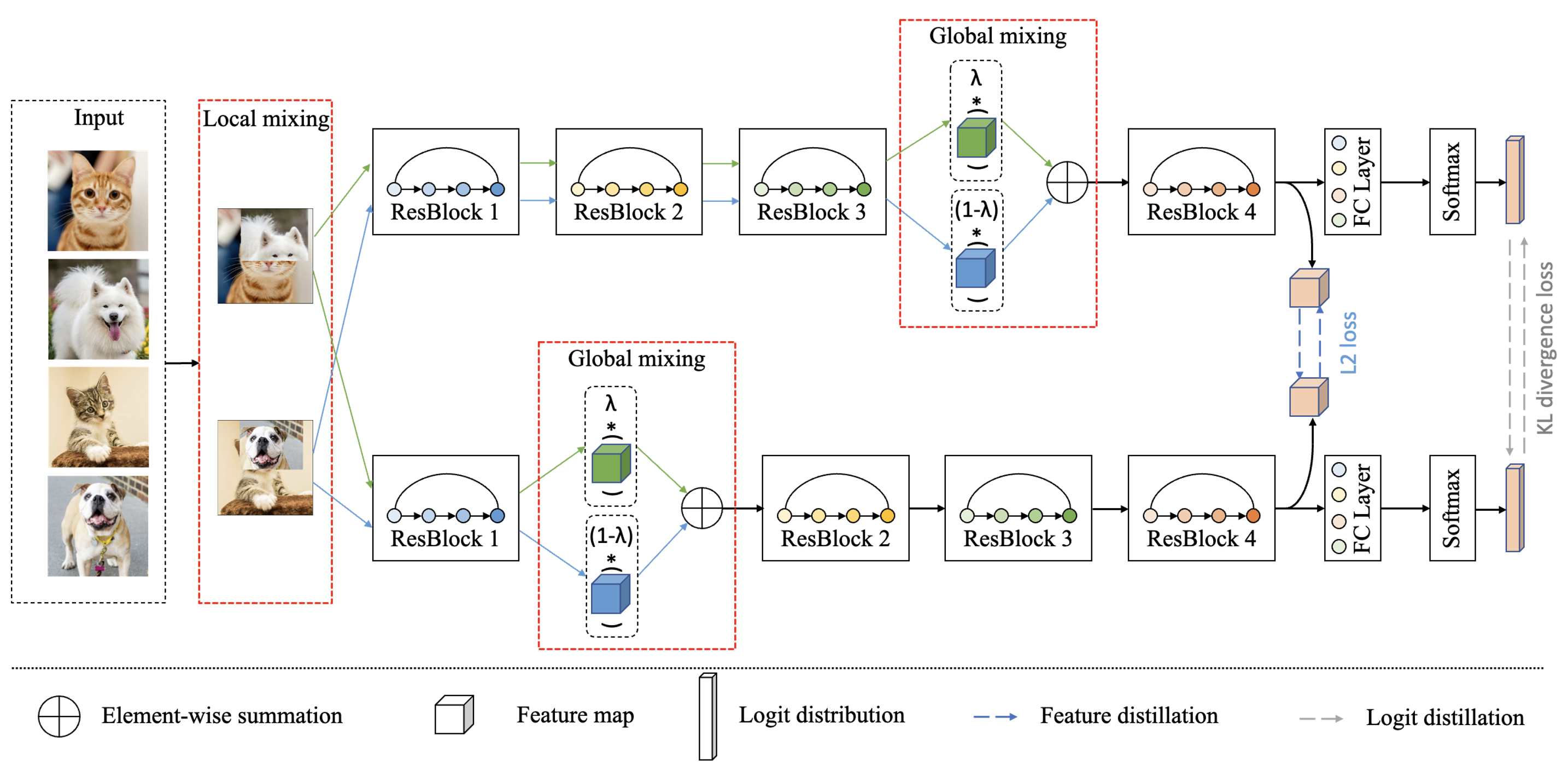}
\end{center}
   \caption{This figure shows a two-student online KD scheme equipped with the proposed MetaMixer. MetaMixer uses a two-stage mixing strategy consisting of local mixing and global mixing, allowing an organized mixing of four different images from two random classes. 
   The local mixing aims to foster localizable low-level knowledge among students and is implemented prior to image input. In contrast, the global mixing encourages  knowledge transferring from different levels of abstraction and can happen at any stage of feature extraction (E.g. Prior to any residual block of ResNet networks).}
\label{fig:overview}
\end{figure*}

\section{Introduction}
    In recent years, we have witnessed the success of deep neural networks in various computer vision tasks~\cite{krizhevsky2017imagenet, he2016deep, girshick2015fast, long2015fully, he2017mask, zhao2017pyramid}. The success of deep learning has been largely driven by ever-larger models~\cite{simonyan2014very, he2016deep}. However, whilst training large models helps to improve the performance, deploying those cumbersome deep models is challenging due to the high computational overhead of such models and their large memory footprint. To address this problem, compression techniques have been investigated. The existing model compression meta-solution can be mainly classified into network pruning~\cite{li2016pruning, molchanov2016pruning, wang2021convolutional}, lightweight structure design~\cite{howard2017mobilenets, li2019learning}, parameter quantization~\cite{han2015deep, wang2019haq}, and knowledge distillation (KD)~\cite{hinton2015distilling}, etc.
    
    As one of the most successful model compression schemes, knowledge distillation (KD) has been proposed and proven to have a strong ability to train a compact student network by mimicking the probability distribution (also known as ``soft targets'') generated by a high-capacity pre-trained teacher network. This process allows the student network to learn the ``dark knowledge''~\cite{hinton2015distilling, yuan2020revisiting} from the teacher network and achieve superior performance compared to training independently from scratch.
    
    After the conventional KD was proposed, there are many works to improve the performance of the student network~\cite{romero2014fitnets, cho2019efficacy, zhao2022decoupled, mirzadeh2020improved, heo2019comprehensive, heo2019knowledge, zagoruyko2016paying}. However, in the conventional KD framework, the teacher model is pre-trained and fixed, and the knowledge can only be transferred from the teacher to the student. Additionally, the high-capacity pre-trained teacher is not always available, which further limits the applicability of the conventional KD method.
    To fill the gap, online distillation is proposed~\cite{zhang2018deep, zhu2018knowledge, chen2020online}, which trains a set of student networks from scratch in a peer-teaching manner. Surprisingly, student networks trained using online distillation methods have shown the potential to compete with, and in some cases, even outperform networks trained with a strong pre-trained teacher network~\cite{zhang2018deep}.  


    However, while most existing collaborative learning methods focus on designing more complex training graphs and using auxiliary branches to improve the distillation of the \textit{high-level} dark knowledge like probability distributions and feature embeddings from the feature extractors~\cite{zhang2018deep, guo2020online, chung2020feature, li2022online, zhu2018knowledge, wu2021peer}, the importance of knowledge from low-level features is greatly underestimated. To this end, we propose MetaMixer, a regularization strategy that can boost the transfer of multi-level knowledge, including both low-level knowledge that contains localizable features and high-level knowledge that are more abstract and focuses on the whole image.
    
    Our main contributions are listed as follows: 
    
    (1) We proposed MetaMixer, a regularization method for online knowledge distillation.

    (2) We stressed the importance of low-level knowledge in online KD, and we proposed a distillation strategy that can take advantage of dark knowledge with different levels of abstraction. 
    
    (3) We evaluated MetaMixer on multiple benchmark datasets and network structures and verified our higher performance over the state-of-the-art online KD methods.

\section{Related Work}
\subsection{Conventional Knowledge Distillation}
    The idea of transferring dark knowledge from the high-capacity teacher model to the compact student model was first proposed in~\cite{bucilua2006model}. However, it did not gain significant attention from researchers until the work by Hinton et al.~\cite{hinton2015distilling}, where the Kullback-Leibler (KL) divergence loss is used to minimize the difference between the probability distribution generated by a student network and the soft targets generated by a pre-trained teacher network. Besides the vanilla KD, researchers have explored more methods to extract more salient dark knowledge from the pre-trained teacher network to the student~\cite{romero2014fitnets, huang2017like, park2019relational, peng2019correlation, tung2019similarity, tian2019contrastive}. For example, Fitnets~\cite{romero2014fitnets} proposed to use intermediate feature representation as a hint to guide the training of the student network. Tian et al.~\cite{tian2019contrastive} first used the contrastive learning mechanism to extract the structural knowledge of the teacher network to supervise the training of the student. 
    
\subsection{Online Knowledge Distillation}
    Whilst the success of conventional KD, its two-stage training process can be time-consuming. Also, the high-performance teacher model may not always be available, which limits the conventional KD in some practical scenarios. To address the problem, online knowledge distillation was proposed. Deep mutual learning~\cite{zhang2018deep} first proposed to train a set of student networks in a peer-teaching manner, showing that online KD methods can train student networks that are competitive with ones trained with conventional KD methods.  ONE~\cite{zhu2018knowledge} firstly applied ensemble learning strategy into online distillation. In ONE, multiple auxiliary branches are applied to the networks. By incorporating auxiliary classifiers and a gate controller, predictions from all branches can be integrated to create a powerful soft target that guides the training of each branch. AFD~\cite{chung2020feature} extracted the knowledge from the intermediate feature representations and used an adversarial training mechanism to further improve the performance of online KD. TSB~\cite{li2022online} proposed temporal accumulators which can stabilize the training process by incorporating predictions during the training procedure. KDCL~\cite{guo2020online} added distortions to the training samples of student networks and designed several effective methods to generate the soft targets containing more dark knowledge. MCL~\cite{yang2022mutual} first introduced the contrastive learning mechanism to the field of online distillation. 

\subsection{Image Mixture}
    Image mixture has been a successful data augmentation strategy to improve the robustness and generalization ability of deep networks. Mixup~\cite{zhang2017mixup} casts a pixel-level interpolation between two images and a linear interpolation of training labels. CutMix~\cite{yun2019cutmix} performs a patch-level mixture which can further improve the localization ability of networks. Manifold Mixup~\cite{verma2019manifold} applied interpolations in the intermediate feature representations. In the field of knowledge distillation, Wang et al.~\cite{wang2020neural} use Mixup to augment the training data to achieve a data-efficient distillation. However, all of the existing image mixture techniques are one-step mixing, and none of the previous research explores the two-step mixing technique in the field of knowledge distillation.

\section{The Proposed Method}
    In this section, we present the overall process of our proposed MetaMixer. As shown in Fig.~\ref{fig:overview}, MetaMixer combines the knowledge of different scales and boosts the knowledge transfer with two mixing stages. For the sake of simplicity, we elaborate the proposed method with two student networks. We will also discuss the generalization to multiple student networks later in this section.

\subsection{MetaMixing}
    We abstract our mixing procedure in MetaMixer as MetaMixing, which consists of two stages: \textbf{local mixing} and \textbf{global mixing}. 

    \textbf{Local mixing} is the first stage of the mixing process in our proposed MetaMixing. As shown in Fig.~\ref{fig:overview}, given four samples $c_1,c_2\in\mathcal{C}$ and $d_1,d_2\in\mathcal{D}$ ($\mathcal{C}$ and $\mathcal{D}$ denote two random classes), their local mixing and label, denoted as $(m_1,y_1)$ and $(m_2,y_2)$, can be expressed as:
    \begin{equation}
    \begin{split}
        m_1 &=M \odot c_1 + (1-M) \odot d_1,\\
        m_2 &=M \odot c_2 + (1-M) \odot d_2, \\
        \{y_1, y_2 \}&= \lambda_1 \cdot y_c + (1-\lambda_1) \cdot y_d,
    \end{split}
        \label{eq:input}
    \end{equation}
    where $\lambda_1$ is the combination ratio sampled from a beta distribution $Beta(\alpha_1, \alpha_1)$ and $\odot$ is an element-wise multiplication. $M \in \{0,1\}^{W \times H}$ is a rectangular binary mask with a width of $\sqrt{\lambda_1}\cdot W$ and a height of $\sqrt{\lambda_1}\cdot H$, and the location of the mask $M$ is randomly selected for every local mixing operation.
    It has been proved that this kind of patch-based mixing operation is a robust regularization strategy to train networks with a strong localization capability~\cite{yun2019cutmix}. And we claim that local mixing is helpful in transferring local-aware low-level knowledge between student peers and improving the localization capability of student networks.

    \textbf{Global mixing} is the second stage of MetaMixing implemented in the feature space of networks, as illustrated in Fig.~\ref{fig:overview}. Consider a neural network $f(x) = FC(f_k(g_k(x)))$ consists of a feature extractor$f_k(g_k(\cdot))$ and a linear classifier $FC(\cdot)$, where $g_k(\cdot)$ denotes the first $k$ layers of the network which map the input to the hidden representation at the layer $k$, and $f_k(\cdot)$ denotes the rest part of the feature extractor mapping the hidden representation $g_k(x)$ to the final feature.
    
    Given two random inputs from local mixing $\{m_1,m_2\}$, we can produce a global mixing of feature representation $\tilde{g}_k(m_1, m_2)$ at the $k$-th layer:
    \begin{equation}
        \tilde{g}_k(m_1, m_2) = \lambda_2 \cdot g_k(m_1) + (1-\lambda_2) \cdot g_k(m_2),
        \label{equ:global}
    \end{equation}
    where $\lambda_2$ is the mixing coefficient of a beta distribution $Beta(\alpha_2, \alpha_2)$. Global mixing encourages the student network to transfer knowledge from different levels for an online KD scheme.

    With local mixing and global mixing, the MetaMixer can effectively extract and transfer dark knowledge of different scales between student peers through its two-stage mixing operation and improve the learning ability of the whole model.

\subsection{Formulation of Training Graph}
    In our description, we refer to the network being supervised as the student network, while we name the other networks that guide the training of the student network as the peer-teacher network(s). For online distillation, in a training iteration, the role of the student network and the peer-teacher network is interchanging. For simplicity, we will introduce the loss function with two student networks in detail, and the method can be easily generalized to more networks and will be discussed later in this section. 
    
    Consider an online KD scheme with two networks. We use $f^s$ to denote the student network that is being trained and $f^t$ to denote the peer-teacher network that guides the training of the student network. For either the student network or the peer-teacher network $f^{(\cdot)}$, the forward pass with MetaMixer can be expressed as:
    \begin{equation}
        \tilde{f}^{(\cdot)}(m_1, m_2) = FC^{(\cdot)}(f^{(\cdot)}_k(\tilde{g}^{(\cdot)}_k(m_1, m_2))),
    \end{equation}
    where layer $k$ of global mixing is individually randomly selected with each forward pass. 

    In each training iteration, the role of $f^s$ and $f^t$ is not fixed, and every network in the student cohort is assigned to $f^s$ for once. The supervision of $f^s$ can be expressed as follows.

    \textbf{Classification loss.} In our work, we use cross-entropy as the classification loss. Specifically, for the input $\{(c_1, y_c), (c_2, y_c), (d_1, y_d), (d_2, y_d)\}$ and their local mixing $\{(m_1, y_1),(m_2,y_2)\}$, the loss of student network $f^s$ can be expressed as: 
    \begin{equation}
         \mathcal{L}^s_{cls} = \sum^2_{i=1} (l^s_{cls}(c_i, y_c) + l^s_{cls}(d_i, y_d) + l^s_{cls}(m_i, y_i)),
    \end{equation} 
    where $l^s_{cls}(x,y) = \mathcal{L}_{CE}(\sigma(f^s(x)), y)$, $\sigma(\cdot)$ is the softmax function to normalize the logits to probability distributions, and $\mathcal{L}_{CE}(\cdot)$ is the cross-entropy loss.

    \textbf{Logit KD loss with MetaMixing.} As the soft target of the teacher network is often regarded as a probability distribution that contains dark knowledge~\cite{hinton2015distilling}, we encourage the student network to behave linearly in the logit distribution between the soft targets of the teacher model. Thus, the linear interpolation of feature maps at any depth $k$ (the depth may include 0) should lead to the corresponding linear interpolation of the final probability distributions with the same interpolation proportion. Similar to other KD works, we use Kullback-Leibler (KL) divergence to minimize the difference between the soft targets and the probability distribution generated by the student network. 
    
    Given two inputs $\{m_1, m_2\}$, the logit-based distillation loss for student network $f^s$ can be expressed as:
    
    \begin{equation}
    \resizebox{\linewidth}{!}{
    $
        \mathcal{L}^s_{m\_logit} = \mathcal{T}^2||\sigma(\tilde{f}^s(m1, m2)/\mathcal{T}), \sigma(\tilde{f}^t(m1, m2)/\mathcal{T})||_{KL},
    $
        \label{eq:mlogit}
    }
    \end{equation}
    
    where $\mathcal{T}$ is the temperature parameter and $||\cdot||_{KL}$ stands for KL divergence.

    \textbf{Feature map KD loss.} Beyond the logit distribution, the feature maps often contain valuable knowledge from the models. Similar to the logit distribution, we also encourage the student network to behave consistently in the feature domain. Inspired by FitNet~\cite{romero2014fitnets}, we also use the $l_2$ norm to minimize the difference between the feature maps. Also, to reduce the computational cost, as shown in Fig.~\ref{fig:overview}, we only use the feature map from the last layer of the feature extractor.
    
    For two inputs $\{m_1, m_2\}$ from local mixing, the feature level distillation loss of student network $f^s$ can be expressed as:
    \begin{equation}
        \mathcal{L}^s_{fea} = \frac{1}{HWC}||f^s_k(\tilde{g}^s_k(m_1, m_2)),f^t_k(\tilde{g}^t_k(m_1, m_2))||_2,
    \end{equation}
    where $||\cdot||_2$ is the $l_2$ norm, and $H, W,$ and $C$ are the height, width, and number of channels of the feature maps.

    In our work, since the student network $f^s$ and peer-teacher network $f^t$ are not necessarily identical, there is a case where the feature embeddings of $f^s$ and $f^t$ are of different shapes. Under such conditions, we linearly interpolate the feature embeddings of the peer-teacher network to the shape of the student network to achieve the feature map distillation. 
    
    \textbf{Logit KD loss without global mixing.} While the MetaMixer actively regularizes the probability distribution of samples interpolated between $m_1$ and $m_2$, it is also important to regularize the interpolation endpoints $m_1$ and $m_2$. Thus, we employ a logit distillation on the interpolation endpoints $(m_1, y_1)$ and $(m_2, y_2)$. Moreover, we apply a weight factor $\mathcal{W}$ as a confidence score to guide the distillation process. We measure the confidence of peer models by calculating the cross-entropy of the predictions and the ground truth on the training dataset and give more weight to the student models with a higher confidence score. The confidence score $\mathcal{W}^1_i$ of the student network $f^s$ on sample $(m_i, y_i)$ can be calculated as:
    \begin{equation}
    \begin{aligned}
        \mathcal{W}^s_i = \frac{l^t_{cls}(m_i, y_i)}{l^s_{cls}(m_i, y_i)+l^t_{cls}(m_i, y_i)},
    \end{aligned}
    \end{equation}
    where $l_{cls}(\cdot)$ is the cross-entropy loss between sample $m_i$ and label $y_i$. Moreover, the confidence scores are detached from the training graph to prevent models from collapsing. With the confidence score $\mathcal{W}$, the distillation loss of student network $f^s$ can be expressed as:
    \begin{equation}
    \begin{aligned}
        \mathcal{L}^s_{e\_logit} = &\mathcal{T}^2 \mathcal{W}^s_1 \cdot ||\sigma(f^s(m_1)/\mathcal{T}), \sigma(f^t(m_1)/\mathcal{T})||_{KL} + 
        \\ & \mathcal{T}^2 \mathcal{W}^s_2 \cdot ||\sigma(f^s(m_2)/\mathcal{T}), \sigma(f^t(m_2)/\mathcal{T})||_{KL},
    \end{aligned}
    \end{equation}
    where $\mathcal{T}$ is the same temperature parameter as in Eq.~\ref{eq:mlogit}.
    
    In summary, the overall loss function of the student network $f^s$ can be expressed as:
    \begin{equation}
        \mathcal{L}^s = \mathcal{L}^s_{cls} + \beta \mathcal{L}^s_{m\_logit} + \gamma \mathcal{L}^s_{fea} + \delta \mathcal{L}^s_{e\_logit},
    \end{equation}
    where $\beta, \gamma, \delta$ are balancing hyperparameters. With our hyperparameter search, we find that $\beta=4, \delta=2$ gives the best performance. For the value of $\gamma$, we find that the $\gamma \in (10^{-1}, 10^{-2})$ functions well, and we set the $\gamma=4\times10^{-2}$ to make the magnitude of feature $l_2$ loss comparable with other loss terms.

    \textbf{Generalize to multiple student networks.} The generalization of our method to more student networks is quite straightforward, and it can be regarded as ensembling peer-teacher networks into a single stronger peer-teacher network by averaging the predictions and feature embedding of peer-teacher networks. For example, consider a distillation with $n$ networks, which can be denoted as a student network $f^s$ and $n-1$ peer-teacher networks $\{f^{t1},f^{t2},\cdots,f^{t(n-1)}\}$. We can ensemble all peer-teacher networks $f^{ti}$ into a single strong peer-teacher network $f^t$ by:
    \begin{equation}
    \begin{aligned}
        \tilde{f}^t(x_1, x_2) &= \frac{1}{n-1}\sum_{i=1}^{n-1} \tilde{f}^{ti}(x_1, x_2), \\
        f^t_k(\tilde{g}^t_k(m_1, m_2)) &= \frac{1}{n-1}\sum_{i=1}^{n-1} f^{ti}_k(\tilde{g}^{ti}_k(m_1, m_2)),
    \end{aligned}
    \end{equation}
    and apply the strong peer-teacher network $f^t$ to the loss functions above.

\begin{table*}
\begin{center}
\resizebox{\textwidth}{!}{
\begin{tabular}{cccccccccccc}
\toprule
\multirow{2}{*}{Network} & \multirow{2}{*}{Baseline} & \multicolumn{2}{c}{DML} & \multicolumn{2}{c}{ONE} & \multicolumn{2}{c}{KDCL} & \multicolumn{2}{c}{MCL} & \multicolumn{2}{c}{MetaMixer (Ours)} \\
\cmidrule(lr){3-4}
\cmidrule(lr){5-6}
\cmidrule(lr){7-8}
\cmidrule(lr){9-10}
\cmidrule(lr){11-12}
{} & {}& Avg. & Ens. & Avg. & Ens. & Avg. & Ens. & Avg. & Ens. & Avg. & Ens. \\
\midrule
ResNet-32 &  93.46  &  93.85  & 94.31  &  94.04  & 94.11 &  93.67  & 94.28  &  \underline{94.13} & \underline{94.74}   & \textbf{94.92} & \textbf{95.37}  \\
ResNet-56 &  94.08  &  94.25  & 94.80  &  \underline{94.72}  & 95.01 &  94.35  & 94.86  & 94.62  & \underline{95.06}   & \textbf{95.63} & \textbf{96.09} \\
ResNet-110&  94.47  &  94.61  & 95.00  &  \underline{95.14}  & \underline{95.45} &  94.39  & 94.93  & 94.86  & 95.37   & \textbf{96.07} & \textbf{96.44} \\
WRN-16-2  &  93.79  &  94.14  & 94.51  &  94.35  & \underline{94.74} &  \underline{94.45}  & 94.66  & 94.27  & 94.59   & \textbf{95.27} & \textbf{95.56}  \\
WRN-40-2  &  95.07  &  95.08  & 95.58  &  \underline{95.61}  & \underline{96.02} &  95.22  & 95.73  &  95.27 & 95.65   & \textbf{96.60} & \textbf{96.92}\\
VGG-13-BN &  94.23  &  94.63  & 95.09  &  \underline{94.74}  & 95.07 &  94.60  & 95.05  & 94.68 & \underline{95.11}  & \textbf{95.57} & \textbf{95.94}  \\
ShuffleNetV2 $\times$1 & 92.29 &  93.41  & 94.00  & 93.28  & 93.73 &  \underline{93.84}  & \underline{94.40}  &  93.07 & 93.61   & \textbf{93.96}& \textbf{94.55}\\
\bottomrule
\end{tabular}
}
\resizebox{\textwidth}{!}{
\begin{tabular}{cccccccccccc}
\toprule
\multirow{2}{*}{Network} & \multirow{2}{*}{Baseline} & \multicolumn{2}{c}{DML} & \multicolumn{2}{c}{ONE} & \multicolumn{2}{c}{KDCL} & \multicolumn{2}{c}{MCL} & \multicolumn{2}{c}{MetaMixer (Ours)} \\
\cmidrule(lr){3-4}
\cmidrule(lr){5-6}
\cmidrule(lr){7-8}
\cmidrule(lr){9-10}
\cmidrule(lr){11-12}
{} & {}& Avg. & Ens. & Avg. & Ens. & Avg. & Ens. & Avg. & Ens. & Avg. & Ens. \\
\midrule
ResNet-32 &  71.10  & 72.64   & 74.60  &  72.67  & 74.51 &  72.58  & 74.87  &  \underline{72.70} & \underline{75.23}   & \textbf{74.79} & \textbf{76.45}  \\
ResNet-56 &  72.88  & 74.13   & 75.99  &  \underline{76.51}  & 77.23 &  74.86  & 76.43  &  74.49 & \underline{77.25}   & \textbf{76.93} & \textbf{78.70}    \\
ResNet-110&  74.14  & 75.42   & 77.10  &  \underline{77.97}  & \underline{78.51} &  75.76  & 77.72  &  76.20 & 78.26   & \textbf{78.58} & \textbf{80.50}    \\
WRN-16-2  &  72.96  & 74.07   & 75.68  &  74.05  & 75.04 &  \underline{75.18}  & \underline{76.90}  &  74.51 & 76.73   & \textbf{76.40} & \textbf{77.88}    \\
WRN-40-2  &  75.87  & 77.35   & 79.15  &  \underline{79.04}  & \underline{80.23} &  78.02  & 79.85  &  77.55 & 79.69   & \textbf{80.18} & \textbf{81.91}    \\
VGG-13-BN &  74.08  & 75.74   & 77.12  &  75.93  & 77.06 &  77.01  & \underline{78.50}  &  \underline{77.02} & 78.42   & \textbf{78.28} & \textbf{79.47} \\
ShuffleNetV2 $\times$1& 72.52   & 74.66  & 76.34  & 74.31 & 75.57 & \underline{74.84}  & \underline{77.20} & 73.21 & 75.84 & \textbf{76.96} & \textbf{78.20}  \\
\bottomrule
\end{tabular}
}
\end{center}
\caption{Top-1 test accuracy (\%) compared with various online distillation methods on CIFAR-10 (top) and CIFAR-100 (bottom) dataset by training two networks with the same architecture. The baseline refers to the performance of the network when it is trained independently with only cross-entropy loss. Due to the special configuration of ONE, we regard the different branches as different networks. The \textit{Avg.} denotes the average performance of the models and \textit{Ens.} refers to the ensemble performance of two networks. Reported the average performance of three runs.}
\label{table:cifar-100}
\end{table*}
\begin{table*}[t]
\begin{center}
\resizebox{\textwidth}{!}{
\begin{tabular}{ccccccccccccc}
\toprule
\multicolumn{2}{c}{Network} & \multicolumn{2}{c}{Baseline} & \multicolumn{3}{c}{DML} & \multicolumn{3}{c}{KDCL} & \multicolumn{3}{c}{MetaMixer (Ours)} \\
\cmidrule(lr){1-2}
\cmidrule(lr){3-4}
\cmidrule(lr){5-7}
\cmidrule(lr){8-10}
\cmidrule(lr){11-13}
Net 1 & Net 2 & Net 1 & Net 2 & Net 1 & Net 2 & Ens. & Net 1 & Net 2 & Ens. & Net 1 & Net 2 & Ens.  \\
\midrule
ResNet-20 & ResNet-32 &  92.76 & 93.46 & 92.91 & 93.50 & 93.87 & \underline{93.16}& \underline{93.82}& \underline{94.04}& \textbf{94.16} & \textbf{94.96} & \textbf{95.24}\\
ResNet-56 & ResNet-110&  94.08 & 94.47 & \underline{94.39} & \underline{94.88} & \underline{95.14} & 94.24& 94.49& 94.90& \textbf{95.76} & \textbf{96.19} & \textbf{96.40}\\
ResNet-32 & WRN-16-2  &  93.46 & 93.79 & 93.74 & 94.15 & 94.58 & \underline{94.21}& \underline{94.55}& \underline{94.81}& \textbf{95.14} & \textbf{95.17} & \textbf{95.59}\\
WRN-16-2 & WRN-28-2   &  93.79 & 94.94 & 94.47 & 95.05 & 95.27 & \underline{94.61}& \underline{95.43}& \underline{95.63}& \textbf{95.27} & \textbf{96.23} & \textbf{96.29}\\
WRN-16-2 & WRN-40-2   &  93.79 & 95.15 & 94.33 & 95.27 & 95.30 & \underline{94.53}& \underline{95.55}& \underline{95.78}& \textbf{95.31} & \textbf{96.54} & \textbf{96.44}\\
VGG-13-BN& WRN-16-2   &  94.23 & 93.79 & 94.57 & 94.15 & 95.03 & \underline{94.70} & \underline{94.33} & \underline{95.31} & \textbf{95.91} & \textbf{95.26} & \textbf{96.02}\\
\bottomrule
\end{tabular}
}

\resizebox{\textwidth}{!}{
\begin{tabular}{ccccccccccccc}
\toprule
\multicolumn{2}{c}{Network} & \multicolumn{2}{c}{Baseline} & \multicolumn{3}{c}{DML} & \multicolumn{3}{c}{KDCL} & \multicolumn{3}{c}{MetaMixer (Ours)} \\
\cmidrule(lr){1-2}
\cmidrule(lr){3-4}
\cmidrule(lr){5-7}
\cmidrule(lr){8-10}
\cmidrule(lr){11-13}
Net 1 & Net 2 & Net 1 & Net 2 & Net 1 & Net 2 & Ens. & Net 1 & Net 2 & Ens. & Net 1 & Net 2 & Ens.  \\
\midrule
ResNet-20 & ResNet-32 & 69.38 & 71.10 & 70.37 & 72.49 & 73.52 & \underline{70.60} & \underline{72.84} & \underline{73.96} & \textbf{71.96} & \textbf{74.83} & \textbf{75.17}\\
ResNet-56 & ResNet-110& 72.88 & 74.14 & 74.03 & 75.82 & 76.67 & \underline{74.30} & \underline{75.92} & \underline{77.28} & \textbf{76.67} & \textbf{78.77} & \textbf{79.63}\\
ResNet-32 & WRN-16-2  & 71.10 & 72.96 & 72.83 & 74.09 & 75.38 & \underline{73.44} & \underline{74.74} & \underline{76.22} & \textbf{75.10} & \textbf{76.26} & \textbf{77.19}\\
WRN-16-2  & WRN-28-2  & 72.96 & 75.14 & 74.10 & 76.31 & 76.99 & \underline{75.06} & \underline{77.15} & \underline{78.53} & \textbf{76.21} & \textbf{79.02} & \textbf{79.38}\\
WRN-16-2  & WRN-40-2  & 72.96 & 75.87 & 74.17 & 77.44 & 77.81 & \underline{75.13} & \underline{78.09} & \underline{78.68} & \textbf{76.22} & \textbf{80.04} & \textbf{80.11}\\
VGG-13-BN & WRN-16-2  & 74.08 & 72.96 & 75.65 & 73.99 & 77.21 & \underline{77.21} & \underline{74.98} & \textbf{79.12} & \textbf{78.76} & \textbf{75.80} & \underline{78.87}\\
\bottomrule
\end{tabular}
}
\end{center}
\caption{Top-1 test accuracy (\%) compared with various online distillation methods on CIFAR-10 (top) and CIFAR-100 (bottom) dataset by training two networks with different architectures. All the notations are the same as the ones in Table.~\ref{table:cifar-100}.}
\label{table:diffarch}
\end{table*}

\begin{table*}
\begin{center}

\begin{tabular}{ccccccc}
\toprule
Network & Baseline & DML & ONE & KDCL & MCL & MetaMixer (Ours) \\
\midrule
ResNet-18 & 69.72 & 69.88 & 70.14 & 69.83 & \underline{70.32}  & \textbf{70.88}  \\
\bottomrule
\end{tabular}

\end{center}
\caption{Top-1 test accuracy (\%) compared with various online distillation methods on ImageNet by training two networks with the same architecture. The result of MCL is from the original paper~\cite{yang2022mutual}. Reported the average performance of 3 runs.}
\label{table:imagenet}
\end{table*}

\section{Experiments and Results}
\subsection{Experimental Setup}
    In this section, we evaluate our proposed method with benchmark network structures including ResNet~\cite{he2016deep}, Wide ResNet (WRN)~\cite{zagoruyko2016wide}, VGG~\cite{simonyan2014very}, and ShuffleNetV2~\cite{ma2018shufflenet} on benchmark datasets including CIFAR-10/100~\cite{krizhevsky2009learning} and ImageNet~\cite{deng2009imagenet}. For the VGG experiments, we use VGG models with batch normalization. For the efficiency of distillation, instead of training from scratch, we separately pre-trained the student networks using cross-entropy loss on the target dataset as the baseline and transferred the weights for the online distillation.

    For CIFAR-10/100 experiments, we train each network for 160 epochs with a batch size of 128. We used stochastic gradient descent (SGD) optimizers with a momentum of 0.9 and a weight decay rate of $5\times10^{-4}$. 
    The initial learning rate of optimizers is set to 0.05, and it is multiplied by 0.1 at the 40th, 70th, 100th, and 130th epoch. We warm up all distillation loss linearly from 0 for 20 epochs. The temperature of logit distillation is set to be 4, while we find that a temperature of 3 also works well.

    For ImageNet experiments, we train each student network for 100 epochs with a batch size of 256. We also use SGD optimizers with a momentum of 0.9 and a weight decay of $1\times10^{-4}$. The initial learning rate is set to 0.02, and it is divided by 10 at the 10th, 40th, and 70th epoch.
    For the training data augmentation, we follow the standard data augmentation strategy as described in~\cite{krizhevsky2017imagenet}. For the test phase, images are resized to $256\times256$ and then center-cropped to $224\times224$. The temperature of logit distillation is set to be 4. 
    
    We compared our proposed method with several state-of-the-art online KD methods, including DML~\cite{zhang2018deep}, ONE~\cite{zhu2018knowledge}, KDCL~\cite{guo2020online},and MCL~\cite{yang2022mutual}. As ONE is configured to share low-level layers for different network branches, in our experiment, we treat different branches of ONE as different networks. For a fair comparison, all of the experimental results are the average over three runs, and we rounded the results to two decimal places. Besides measuring the performance of each student network, we also measure the ensemble performance of the student models by averaging their predictions.
    
\begin{table*}
\begin{center}
\begin{tabular}{cccccccccc}
\toprule
\multirow{2}{*}{Network} & \multirow{2}{*}{Baseline} & \multicolumn{2}{c}{Without mixing} & \multicolumn{2}{c}{Global Only} & \multicolumn{2}{c}{Local Only} & \multicolumn{2}{c}{MetaMixer} \\
\cmidrule(lr){3-4}
\cmidrule(lr){5-6}
\cmidrule(lr){7-8}
\cmidrule(lr){9-10}
{} & {} & Avg. & Ens.& Avg. & Ens.& Avg. & Ens. & Avg. & Ens. \\
\midrule
ResNet-32 & 71.10 &  72.91 & 74.86 &73.68   &  75.39    & 74.54  &  76.43    & 74.79 & 76.45\\
ResNet-56 & 72.88 &  74.75 & 76.83 &75.21   &  77.38    & 76.72  &  78.62    & 76.93 & 78.70\\
WRN-16-2  & 72.96 &  74.52 & 76.25 &75.14   &  76.73    & 76.09  &  77.94    & 76.40 & 77.88\\
\bottomrule
\end{tabular}
\end{center}
\caption{Top-1 test accuracy (\%) comparison on CIFAR-100 by mutually training two networks of the same structure with different mixing stages of MetaMixer.}
\label{table:stage}
\end{table*} 


\begin{figure*}
\begin{center}
    \subfloat[ResNet Performance]{
       \includegraphics[height=0.16\textwidth]{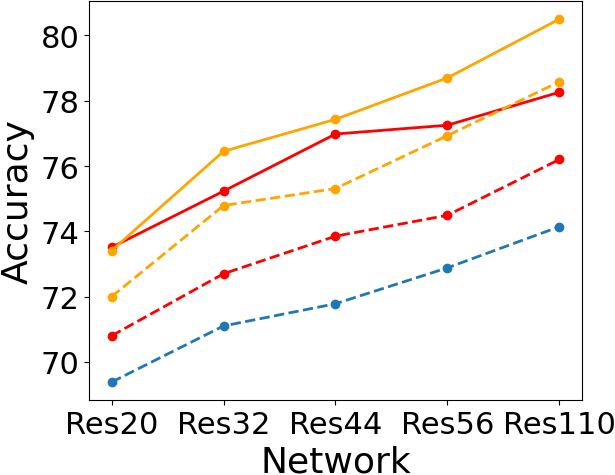}}
    \label{2a}\hfill
    \subfloat[ResNet Performance Gain]{
        \includegraphics[height=0.16\textwidth]{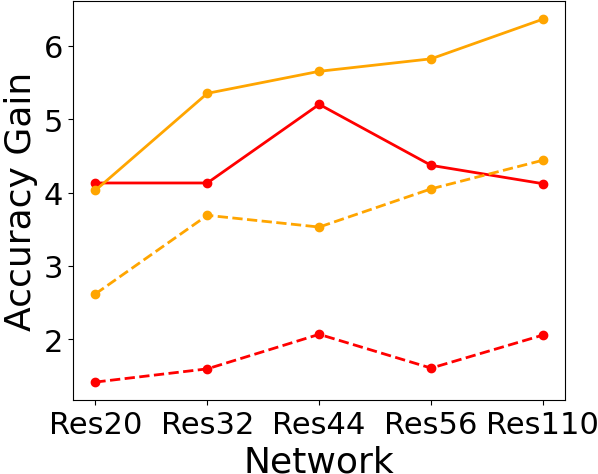}}
    \label{2b}\hfill    
    \subfloat[WRN Performance]{
       \includegraphics[height=0.16\textwidth]{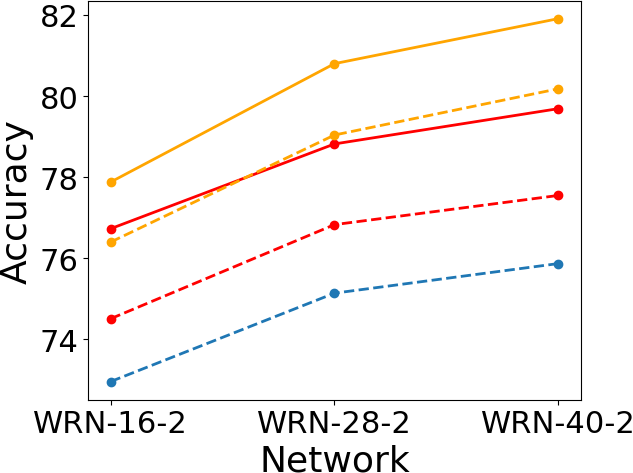}}
    \label{2c}\hfill
    \subfloat[WRN Performance Gain]{
        \includegraphics[height=0.16\textwidth]{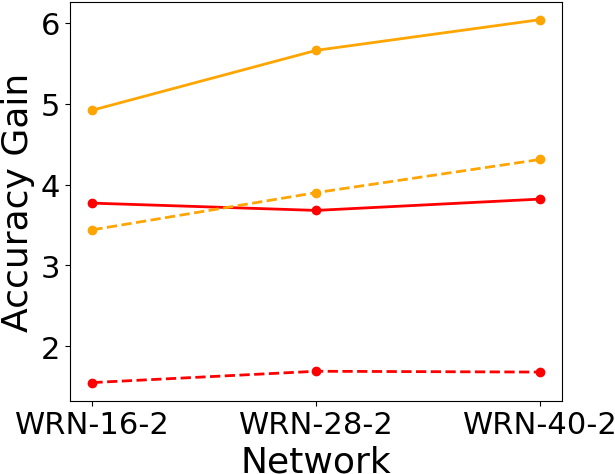}}
    \label{2d}\hfill
    \captionsetup[subfigure]{labelformat=empty}
    \subfloat[]{%
    {
    \raisebox{0.05\textwidth}{\includegraphics[height=0.08\textwidth]{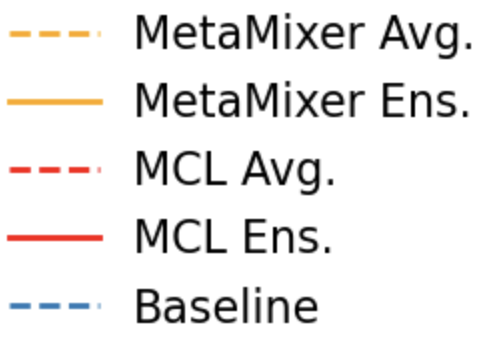}}
    }
    }      
    \caption{The performance and performance gain of mutually training two ResNet or Wide ResNet models with the same structure of different model depth and network capacity on CIFAR-100 dataset. From the result, we can see that compared with MCL which tends to have a fixed performance gain, the performance gain of our method increases steadily with the increase of network depth and capacity.}
    \label{fig:depth} 
\end{center}
\end{figure*}

\subsection{Main Results}
\subsubsection{Results on CIFAR-10/100}

\label{subsubsec:cifar-result}
    We mainly evaluated the performance of our proposed MetaMixer on CIFAR-10/100 dataset with two student networks, while the performance with more student networks is available in our ablation study in section~\ref{sec:multi}. For MetaMixer, since the network structure of student networks is not necessarily identical, we classified our experiments into two categories: student networks with the same structure and student networks with different structures. For the experiments where the student models are of different architectures, we tried several representative combinations of models, including models of different depth/capacity, and models of different types.

    \textbf{Same architecture.} From the experimental results in Table~\ref{table:cifar-100}, we find that MetaMixer consistently outperforms the state-of-the-art online KD methods on both CIFAR-10 and CIFAR-100 in terms of average performance and ensemble performance. 
    Besides the improved performance, on CIFAR-100, we find an interesting phenomenon that compared with other methods which tend to have a fixed performance gain over the baseline, the performance gain of MetaMixer increases steadily with deeper and larger models, and we will investigate this feature in the ablation study in section~\ref{sec:depth}. 

    \textbf{Different architectures.} In this experiment, we validate the capability of different methods in dealing with model peers of different structures and model capacities. The results of ONE and MCL are not presented for models with different architectures. This is because ONE cannot be applied when two networks are of different architectures due to the configuration of sharing the low-level layers. Also, as MCL takes the advantage of contrastive learning mechanism, the performance of MCL gets greatly negatively affected with models of different capacities.

    As indicated in Table~\ref{table:diffarch}, when the models are of different architectures, the performance advantage of MetaMixer over other online KD approaches is still significant both in terms of the average performance and the ensemble performance.

\subsubsection{Results on ImageNet}
    We also evaluated the performance of MetaMixer on the ImageNet dataset which is much larger than CIFAR-10/100. We trained two peer networks of the same structure mutually with different methods and analyzed the mean performance of the two networks, as shown in Table~\ref{table:imagenet}. Our experimental result shows that the MetaMixer outperforms all of the online KD methods, which validates the effectiveness of our methods on the large-scale dataset.
    
\subsection{Ablation Studies and Analyses}
\subsubsection{Effect of Different Mixing Stages in MetaMixer}
    Our proposed MetaMixer contains two consecutive mixing stages, which are local mixing and global mixing. In this ablation study, we investigate the effect of different mixing stages by training two networks with different mixing stages on CIFAR-100. From the experimental result in Table~\ref{table:stage}, we can see that either global mixing or local mixing can improve the average and ensemble performance, and the combination of local and global mixing can provide further performance gain in terms of average accuracy. Interestingly, during this ablation study, we found that the ensemble performance of networks trained only with the local mixing can be sometimes comparable with the ensemble performance of MetaMixer, leaving it an interesting point worth further investigating.

\begin{table}
\begin{center}
\resizebox{\linewidth}{!}{
\begin{tabular}{ccccccccc}
\toprule
\multirow{2}{*}{Network} & \multicolumn{2}{c}{2 models} & \multicolumn{2}{c}{3 models} & \multicolumn{2}{c}{4 models} \\
\cmidrule(lr){2-3}
\cmidrule(lr){4-5}
\cmidrule(lr){6-7}
{} & Avg. & Ens.& Avg. & Ens.& Avg. & Ens. \\
\midrule
Res-32 & 74.79 & 76.45 & 74.91 & 77.69 & 74.92& 77.54\\
Res-56 & 76.93 & 78.70 & 76.93 & 79.66 & 76.91& 79.64\\
W-16-2  & 76.40 & 77.88 & 76.48 & 79.02& 76.39& 78.66\\
\bottomrule
\end{tabular}}
\end{center}
\caption{Top-1 test accuracy (\%) of mutually training multiple networks with the same architecture on CIFAR-100 dataset using MetaMixer.}
\label{table:multi}
\end{table}

\subsubsection{Performance with Multiple Student Networks}
\label{sec:multi}
    To further evaluate the effectiveness of our proposed method, we generalized our online KD method to multiple student networks to evaluate the performance. Specifically, we jointly trained multiple networks with the same architecture on CIFAR-100 dataset and evaluated the average performance along with the ensemble performance. As shown in Table~\ref{table:multi}, when the size of the student cohort increases from two to three, the average performance improvement is small compared with the two-model case. However, we witness a significant improvement in the ensemble performance. 
    Also, we find that both the average and the ensemble performance saturate at three models.


\subsubsection{Performance Gain vs. Network Capacity}
\label{sec:depth}
    Experiments were also conducted to evaluate the performance gain of our method with the increase of network depth and capacity. In this experiment, we use ResNet and Wide ResNet series on the CIFAR-100 dataset. We jointly train two networks with the same architecture and compared the performance gain of MetaMixer with other different online KD methods. As shown in Fig.~\ref{fig:depth}, different from MCL which tends to have a fixed performance gain, the average and ensemble performance gain of MetaMixer increases steadily with the increase of network depth and capacity. We believe this is a feature helpful for training larger models. We hypothesize that the reason behind this might be that MetaMixer, as a regularization strategy, can augment meaningful samples in the training process to avoid overfitting, while other methods like MCL and KDCL focus on designing better supervision for the student network and have nothing to do with regularization and overfitting issue. With the overfitting issue alleviated, larger models can benefit more from extra parameters and generalize better.
    
\begin{figure}
\begin{center}
    \subfloat[Performance gain when $\alpha_2=0.2$]{
       \includegraphics[width=0.48\linewidth]{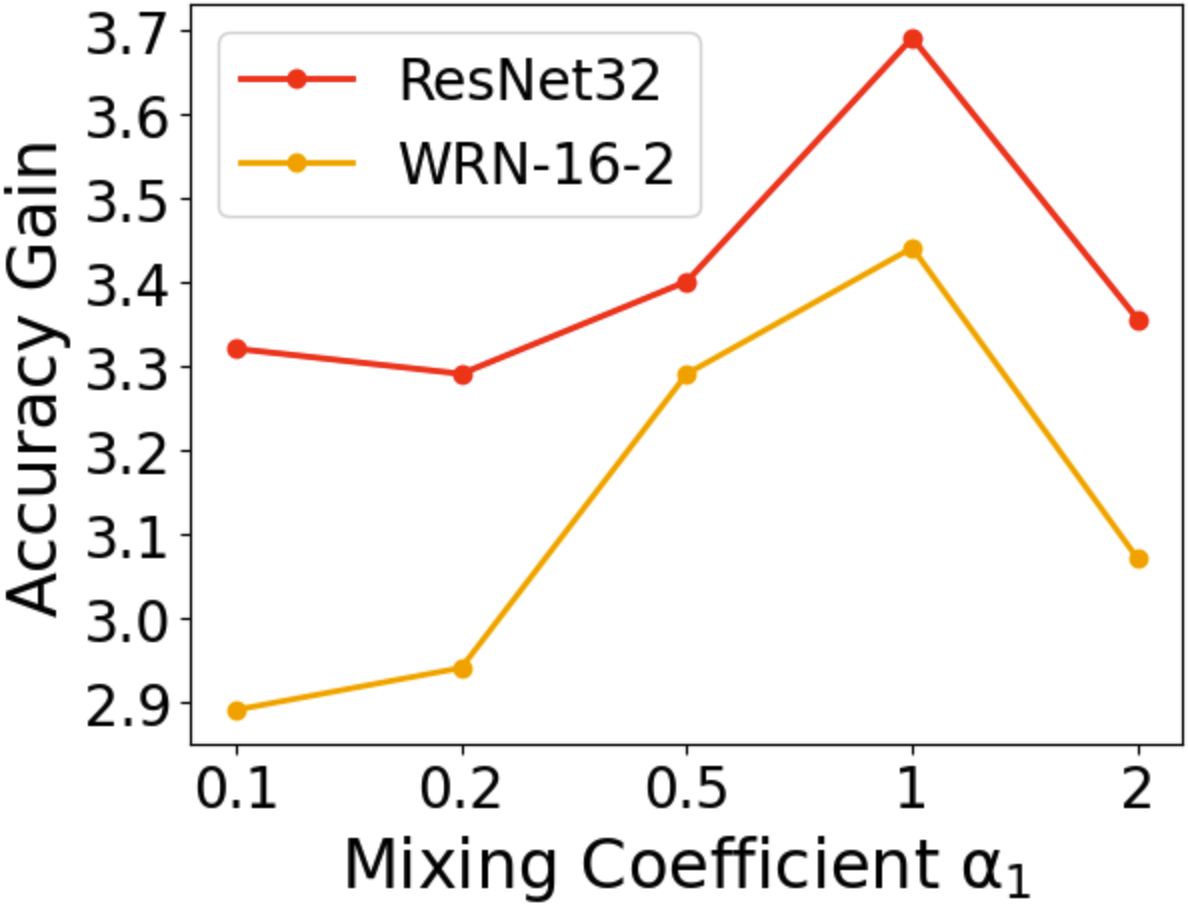}}
    \label{m1}\hfill
    \subfloat[Performance gain when $\alpha_1=1$]{
        \includegraphics[width=0.48\linewidth]{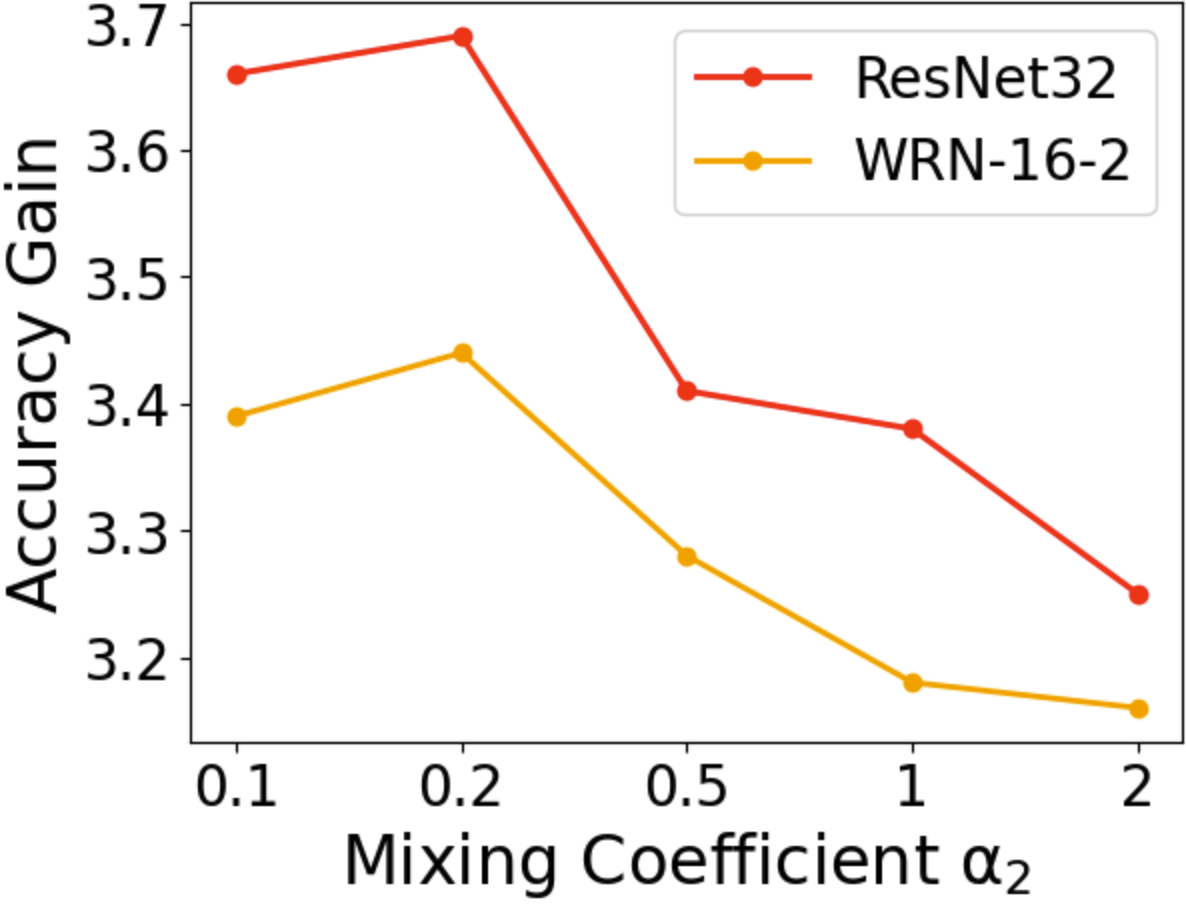}}
    \label{m2}
    \caption{The impact of $\alpha_1$ and $\alpha_2$ on the average performance gain on CIFAR-100. As shown in the figure, the best performance is achieved with $\alpha_1=1$ and $\alpha_2=0.2$.}
    \label{fig:coefficient} 
\end{center}
\end{figure}

\subsubsection{Ablation of Mixing Coefficient $\lambda_1$ and $\lambda_2$} 
    In MetaMixer, there are two different mixing stages: local mixing and global mixing. Thus, two different mixing coefficients $\lambda_1$ and $\lambda_2$ as described in Eq.~\ref{eq:input} and Eq.~\ref{equ:global} are introduced into our method. As indicated above, both of the mixing coefficients are sampled from beta distributions where $\lambda_1\sim Beta(\alpha_1, \alpha_1)$ and $\lambda_2\sim Beta(\alpha_2, \alpha_2)$. In this ablation, we evaluated the performance of MetaMixer with $\alpha_1, \alpha_2 \in \{0.1, 0.2, 0.5, 1, 2\}$ and searched the best combination of $\alpha_1$ and $\alpha_2$ on the CIFAR-100 dataset using two networks with the same structure. We measured the average performance gain under different combinations of $\alpha_1$ and $\alpha_2$ as shown in Fig.~\ref{fig:coefficient}. The best performance is achieved with $\alpha_1 = 1$ and $\alpha_2 = 0.2$.
    
\begin{figure}
\begin{center}
    \subfloat[DML]{
       \includegraphics[width=0.48\linewidth]{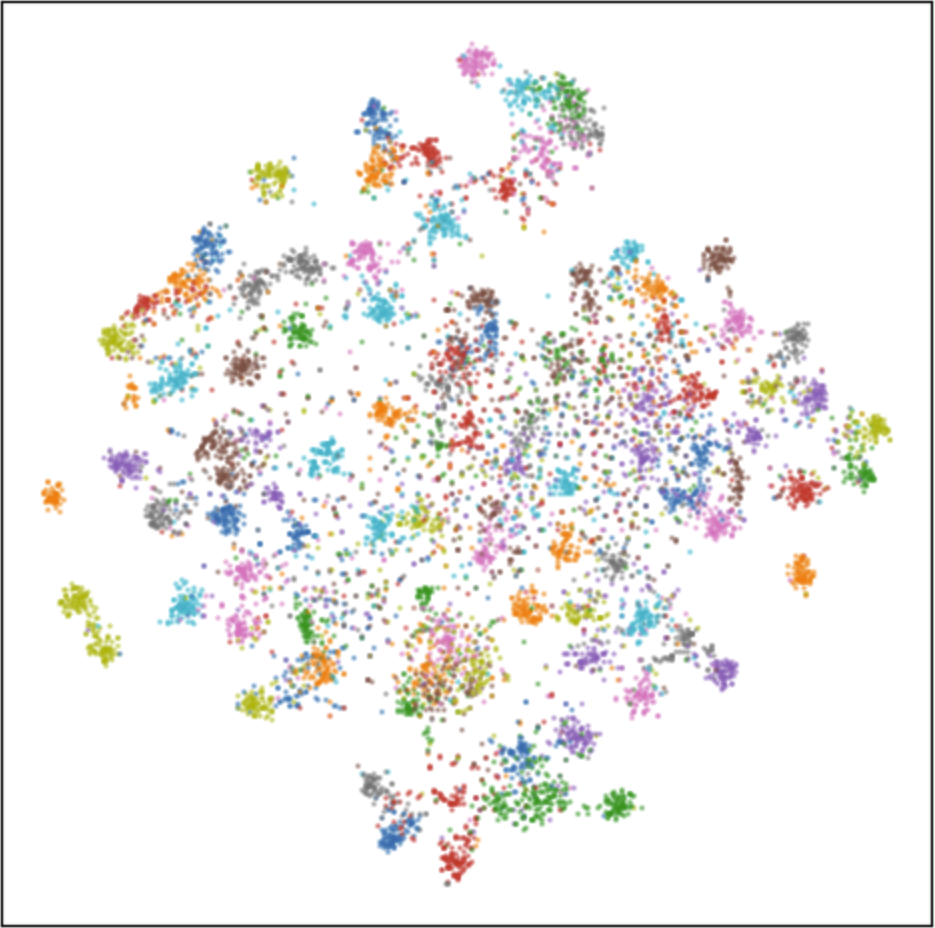}}
    \label{1a}\hfill
    \subfloat[KDCL]{
        \includegraphics[width=0.48\linewidth]{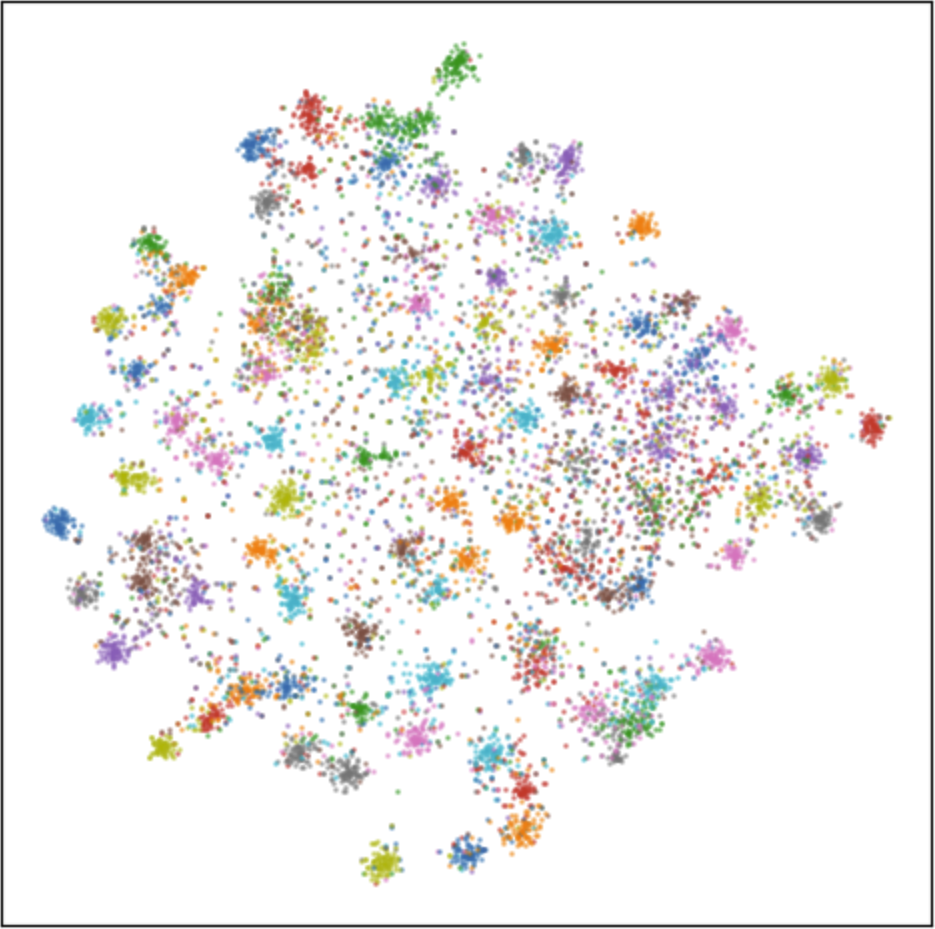}}
    \label{1b}\\
	  \subfloat[MCL]{
        \includegraphics[width=0.48\linewidth]{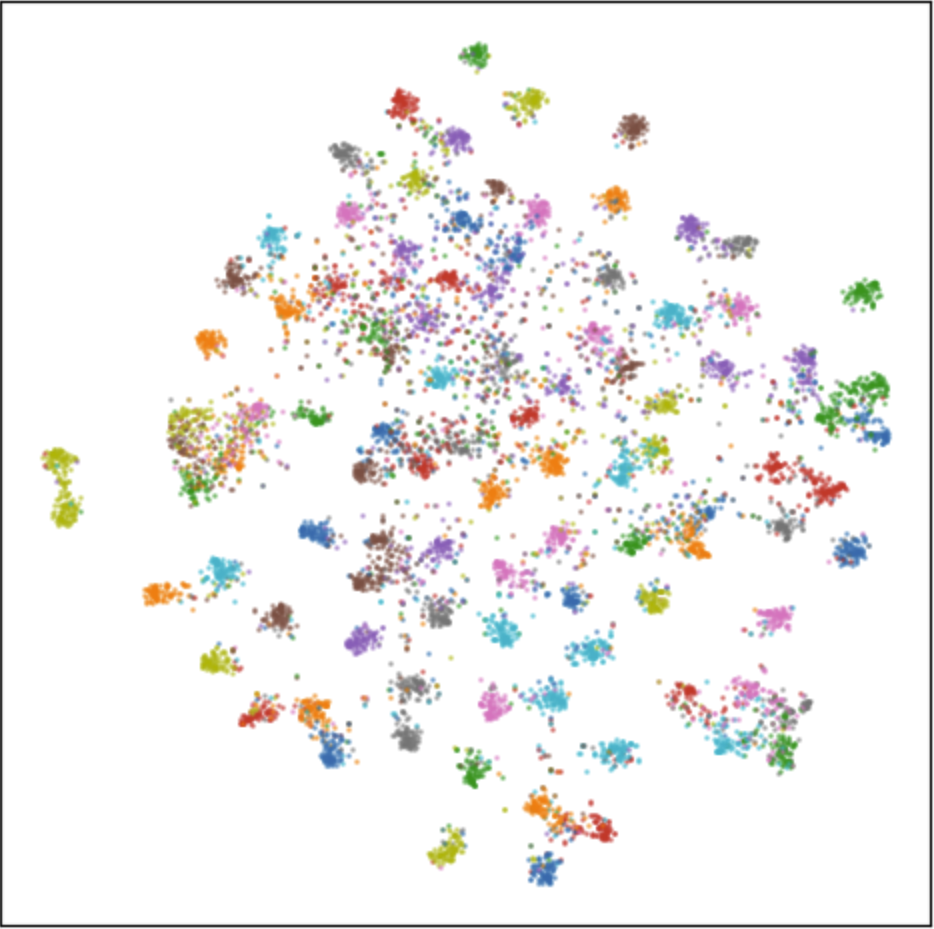}}
    \label{1c}\hfill
	  \subfloat[MetaMixer (Ours)]{
        \includegraphics[width=0.48\linewidth]{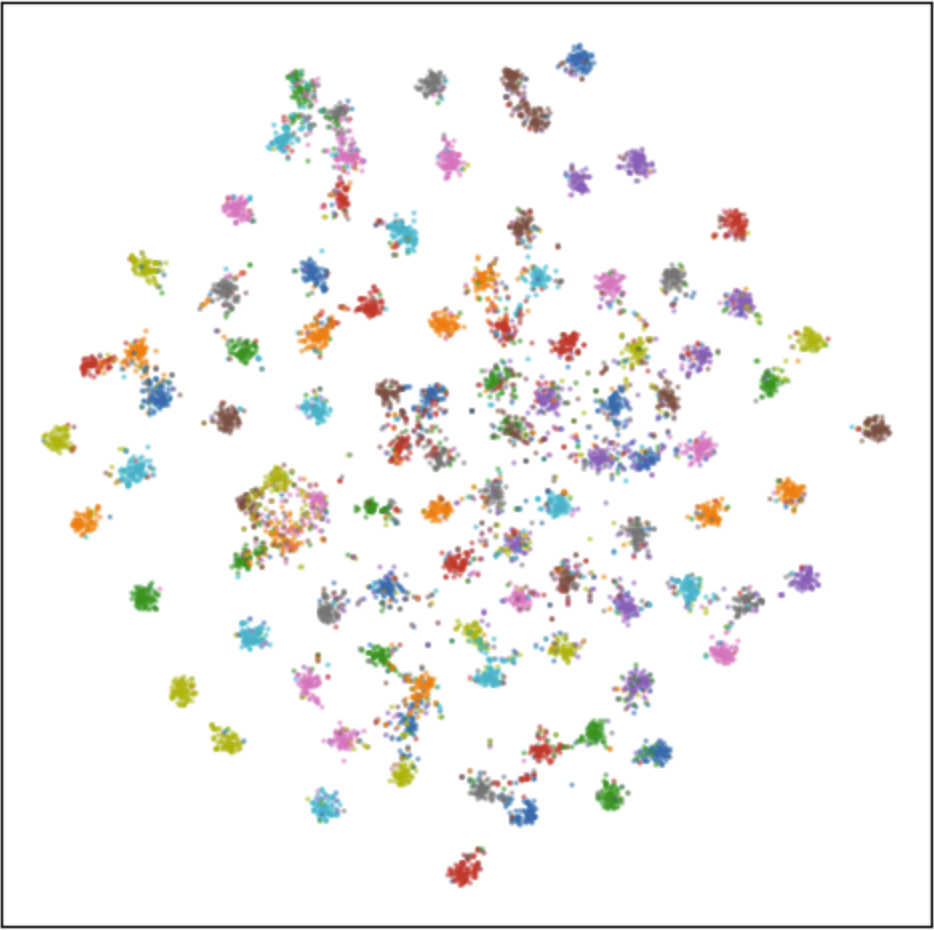}}
    \label{1d} 
    \caption{T-SNE visualization of embedding spaces generated by ResNet-110 models trained with different online KD methods on CIFAR-100. The figure shows that the model trained with MetaMixer can produce feature representations that are more separable compared with models trained with other online KD methods.}
    \label{fig:t-SNE} 
\end{center}
\end{figure}

\subsubsection{T-SNE Visualization} 

    In this experiment, we present the t-SNE~\cite{van2008visualizing} visualization of the embedding space generated by models trained with different online KD methods. Specifically, we mutually trained two ResNet-110 networks on CIFAR-100 with different online KD methods including DML, KDCL, MCL, and MetaMixer. As shown in Fig.~\ref{fig:t-SNE}, the experimental result shows that the embedding space produced by models trained with the MetaMixer is more separable than those trained with other online KD methods. 

\section{Conclusion}
    

    In this paper, we present a novel perspective on online KD through a regularization strategy that involves a two-stage mixing operation.
    More importantly, we highlight the limitation of existing online KD methods, which mainly concentrate on improving the distillation of high-level knowledge such as probability distribution, while neglecting the value of low-level knowledge in collaborative learning. In contrast, MetaMixer integrates knowledge from multiple scales and achieves significant improvements over the state-of-the-art online KD methods on representative benchmark datasets. We hope our work can foster future research on online KD and collaborative learning.

{\small
\bibliographystyle{ieee_fullname}
\bibliography{egbib}
}

\end{document}